\documentclass[letterpaper, 10 pt, conference]{ieeeconf}  

\IEEEoverridecommandlockouts                              

\newcommand{\mypara}[1]{\vspace{2pt}\noindent\textbf{#1}}

\usepackage[labelformat=simple]{subcaption}  

\usepackage[font=small]{caption}
\usepackage{graphicx} 
\usepackage{tabularx}
\usepackage{times} 
\usepackage{amsmath} 
\usepackage{amssymb}  
\usepackage{comment}
\usepackage{url}
\usepackage{multirow}
\usepackage{colortbl}
\usepackage{booktabs}
\usepackage[table,xcdraw]{xcolor}
\usepackage[normalem]{ulem}
\newcolumntype{x}[1]{>{\centering\arraybackslash}p{#1pt}}
\newcolumntype{y}[1]{>{\raggedright\arraybackslash}p{#1pt}}
\newcolumntype{z}[1]{>{\raggedleft\arraybackslash}p{#1pt}}
\usepackage{xcolor}
\usepackage{pifont} 
\usepackage{array}
\usepackage{soul}

\makeatletter
\let\NAT@parse\undefined
\makeatother
\usepackage[pagebackref=false,breaklinks=true,letterpaper=true,colorlinks,bookmarks=false]{hyperref}
\newcommand{\tablestyle}[2]{\setlength{\tabcolsep}{#1}\renewcommand{\arraystretch}{#2}\centering\footnotesize}
\useunder{\uline}{\ul}{}
\useunder{\cellcolor[HTML]{EFEFEF}}{\hl}{}
\newcommand{\fadedtext}[1]{\textcolor{gray}{#1}}
\newcommand{\cmark}{\ding{51}}%
\newcommand{\xmark}{\ding{55}}%

\title{\LARGE \bf
From Cognition to Precognition:  \\
A Future-Aware Framework for Social Navigation
}

\author{Zeying Gong$^{1}$, Tianshuai Hu$^{2}$, Ronghe Qiu$^{1}$ and Junwei Liang$^{1,2*}$%
\thanks{$^{1}$ The Hong Kong University of Science and Technology (Guangzhou). 
{\tt\footnotesize \{zgong313,rqiu683\}@connect.hkust-gz.edu.cn, junweiliang@hkust-gz.edu.cn} * Corresponding author.}%
\thanks{$^{2}$ The Hong Kong University of Science and Technology. {\tt\footnotesize thuaj@connect.ust.hk}}%
}

\begin{document}

\maketitle
\thispagestyle{empty}
\pagestyle{empty}

\begin{abstract}

To navigate safely and efficiently in crowded spaces, robots should not only perceive the current state of the environment but also anticipate future human movements. 
In this paper, we propose a reinforcement learning architecture, namely \textit{Falcon}, to tackle socially-aware navigation by explicitly predicting human trajectories and penalizing actions that block future human paths. 
To facilitate realistic evaluation, we introduce a novel SocialNav benchmark containing two new datasets, Social-HM3D and Social-MP3D. 
This benchmark offers large-scale photo-realistic indoor scenes populated with a reasonable amount of human agents based on scene area size, incorporating natural human movements and trajectory patterns. 
We conduct a detailed experimental analysis with the state-of-the-art learning-based method and two classic rule-based path-planning algorithms on the new benchmark. 
The results demonstrate the importance of future prediction and our method achieves the best task success rate of 55\% while maintaining about 90\% personal space compliance.    
We will release our code and datasets.
Videos of demonstrations can be viewed at \url{https://zeying-gong.github.io/projects/falcon/}.

\end{abstract}

\section{INTRODUCTION}
\noindent \textbf{Social navigation (SocialNav)} refers to autonomous robot adhereing to \textit{social norms} and \textit{social etiquette} while navigating environments shared with humans~\cite{mavrogiannis2023core}. 
This task poses new challenges to visual navigation, as modular approaches relying on pre-built maps struggle in dynamic, human-populated environments where collision avoidance is crucial.

\begin{figure}[t]
    \centering
    \includegraphics[width=0.95\linewidth]{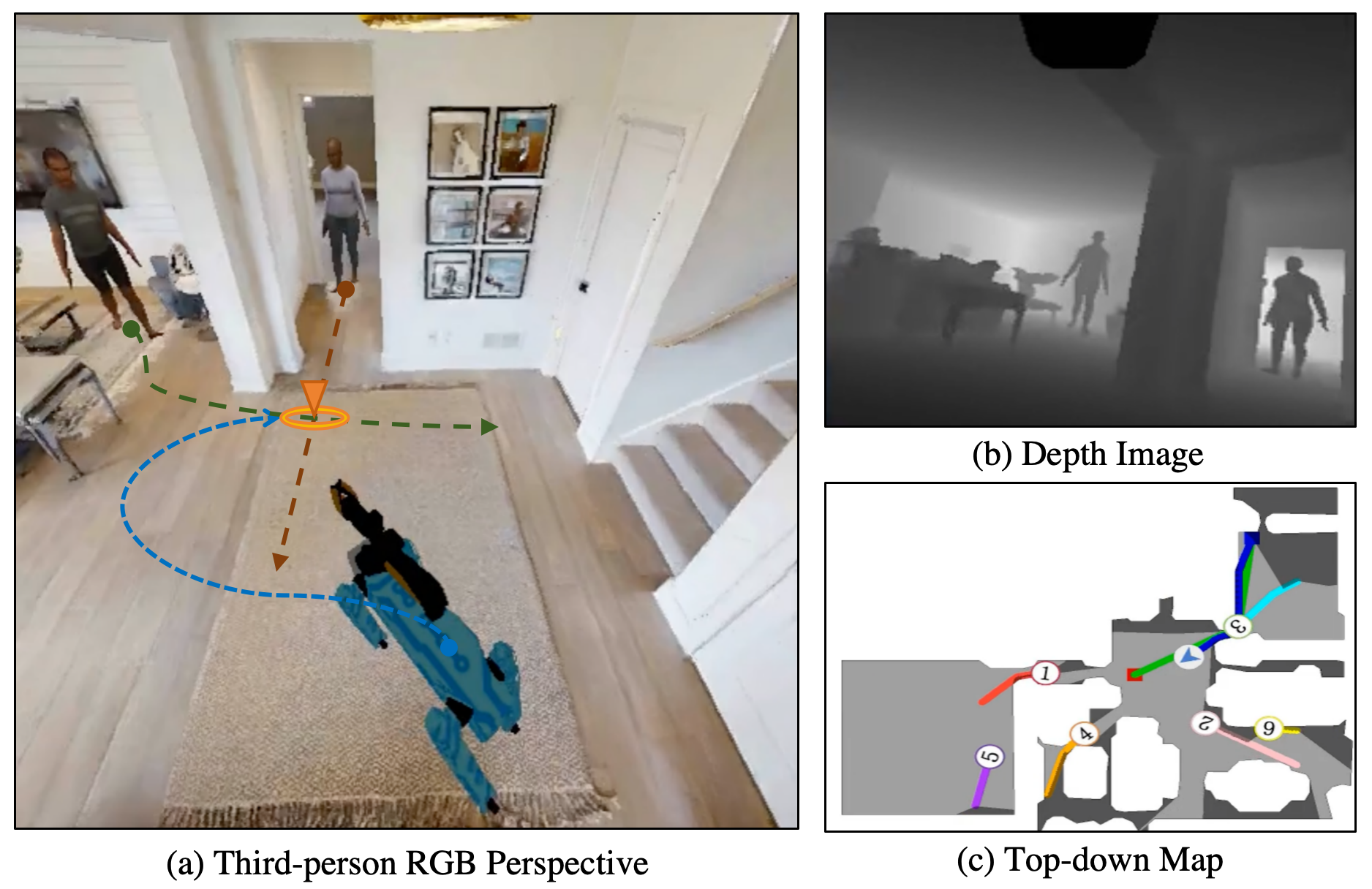}
    \caption{We integrate trajectory prediction into the SocialNav task. In (a), the robot navigates toward a goal while predicting human trajectories (dashed lines) and avoiding them, following social etiquette. The robot uses depth input as shown in (b). (c) offers a top-down map for reference, which is not used by the robot.}
    \label{fig:task_illu}
    \vspace{-2em}
\end{figure}

Previous works typically train navigation agents using reinforcement learning (RL)~\cite{kapoor2023socnavgym,wang2024multi,hirose2024selfi}. 
However, RL-based SocialNav often suffers from short-sighted obstacle avoidance, limiting efficiency around humans~\cite{chen2019crowd,li2019sarl}. A common solution to this problem is hierarchical methods, which combine global planners with low-level RL policies~\cite{perez2021robot}. However, these approaches rely on prior knowledge of the environment for planning~\cite{cancelli2023exploiting}, making them unsuitable for realistic, dynamic scenarios.
Consider the example in Fig.~\ref{fig:task_illu}, where a robot navigates to a goal located at the intersection of two nearby humans' future paths. In such cases, traditional RL approaches may struggle to avoid humans due to limited foresight or reliance on global information.
In contrast, our approach addresses these issues by explicitly predicting human trajectories that enable social compliance and long-term dynamic collision avoidance.

Human trajectory forecasting has been shown to improve collision avoidance and navigation in dynamic environments~\cite{liang2019peeking,nishimura2020risk,mangalam2021goals}, but it is mostly applied in outdoor scenarios like autonomous driving~\cite{liang2020garden,liang2020simaug,zhao2021tnt}. 
This motivates us to integrate trajectory prediction algorithms into the SocialNav task. 
Indoor environments pose additional challenges, such as limited space and maneuverability, increasing the risk of collisions~\cite{perez2021robot}. 
To address these challenges, we develop \textit{\textbf{Falcon}}, a \underline{f}uture-aware Soci\underline{al}Nav framework with pre\underline{co}gnitio\underline{n}.
The key novelties of \textit{Falcon} lie in its future awareness. First, it introduces the \textbf{Social Cognition Penalty}, including a trajectory obstruction penalty, encouraging the agent to proactively avoid potential collisions and follow social etiquette. Second, it employs a \textbf{Spatial-Temporal Precognition Module} that incorporates socially aware auxiliary tasks, including trajectory prediction, to enhance the agent’s comprehension of future dynamics during training. 

Another challenge in SocialNav is the lack of realism in configurations~\cite{francis2023principles}. Current approaches typically simplify the environment to only include a robot and surrounding humans, neglecting the complexity of the scene itself~\cite{liu2021decentralized,trajnetpp,holk2024polite}. 
Moreover, the solutions often assume the robot has access to global information, such as real-time human positions or a full map of the environment~\cite{cancelli2023exploiting}. 
To overcome these unrealistic settings, we introduce a novel SocialNav benchmark that contains two new datasets, \textbf{Social-HM3D} and \textbf{Social-MP3D}.
We construct the datasets using 3D-reconstructed real-world indoor scenes.
The scenes are populated with collision-avoidant humans moving toward their goals.
The robot is given only one point goal and egocentric inputs during inference, without relying on a global map or known human trajectories. 
Our benchmark offers a more realistic representation of social navigation. 
In addition, we verify \textit{Falcon} on the proposed benchmark, achieving state-of-the-art performance with a 55\% success rate while maintaining a high social compliance score.

Overall, our main contributions are as follows:
\begin{itemize}
    \item We introduce the first realistic SocialNav benchmark with two novel datasets, \textbf{Social-HM3D} and \textbf{Social-MP3D}, featuring large-scale photo-realistic scenes with realistic human and robot animations.
    \item We propose a new and effective SocialNav framework, \textit{\textbf{Falcon}}, that integrates explicit trajectory prediction, allowing the robot to perceive and predict for safe, socially comfortable and effective navigation.
    \item We establish a new state-of-the-art result compared to prior approaches on the proposed benchmark.
\end{itemize}

\section{Related Works}
\subsection{Social Navigation.}
In this paper, we focus on the SocialNav task~\cite{perez2021robot,francis2023principles}, which was first introduced in the iGibson SocialNav Challenge~\cite{xia2020interactive}, building on the PointGoal Navigation (PointNav) by adding moving humans.
Humans in the challenge are static figures with unrealistic movements.
In contrast, our work utilizes the recent Habitat 3.0 simulator~\cite{puig2023habitat} to leverage realistic human movements and animations.

SocialNav has been widely studied in robotics, computer vision, and social behavior analysis~\cite{singamaneni2024survey, moller2021survey}. In collision-free multi-agent navigation~\cite{van2011reciprocal, rvo, learned} and dynamic environments~\cite{mpdyn}, research has advanced to address challenges posed by the presence of humans~\cite{guzzi2013human, socialforces, sa-cadrl, social-graph, socialattention}. 
Other approaches~\cite{socialforces, socialattention} model human-agent interactions through spatio-temporal graphs~\cite{social-graph} to capture agent dynamics over time. 
Recent studies have also focused on egocentric SocialNav in photo-realistic or real-world scenarios~\cite{rudenko2020thor, martin2021jrdb, vuong2023habicrowd}. 
Differing from these approaches, our method introduces explicit trajectory prediction within auxiliary tasks to train an RL-based agent for SocialNav.

\subsection{Auxiliary tasks in Navigation.}
Auxiliary tasks can enhance sample efficiency and improve the performance of primary navigation tasks. Mirowski~\cite{mirowski2016learning} introduced self-supervised auxiliary tasks to help agents navigate mazes more efficiently. Since then, a wide range of general auxiliary tasks have been developed~\cite{jaderberg2016reinforcement,lin2019adaptive}, including tasks focused on predicting environmental properties or forecasting the agent’s own states~\cite{agrawal2015learning, jaderberg2016reinforcement, ye2021auxiliary}. These tasks enable agents to better understand and interact within static environment. Recently, auxiliary tasks using privileged information, such as distances or angles between the agent and humans, have been introduced for SocialNav~\cite{cancelli2023exploiting}. 
Our auxiliary tasks not only perceive humans' current positions, but also focus on predicting their future trajectories, facilitating efficient learning of human movement patterns and reducing collisions risks.

\subsection{Human Trajectory Prediction.}
Human trajectory prediction is vital for enabling safe and intelligent behavior in autonomous systems~\cite{rudenko2020human,huang2023multimodal}. Traditional approaches often rely on physical models, such as the Social Force model~\cite{socialforces}, which uses attractive and repulsive forces to simulate social behaviors and collision avoidance. 
Broadly, these methods fall into three categories. One approach is based on physics, where explicit dynamical models are derived from Newton's laws of motion to predict trajectories~\cite{elnagar2001prediction, zernetsch2016trajectory, coscia2018long}. Another set of methods focuses on learning motion patterns from observed historical trajectories~\cite{alahi2016social, kucner2017enabling, vemula2017modeling}. Moreover, planning-based methods aim to reason about the motion intent of rational agents, predicting trajectories by understanding agents' goals and their decision-making processes~\cite{vasquez2016novel, rudenko2017predictive, rehder2018pedestrian}. 
Our approach leverages these insights to develop a method that not only predicts human trajectories but also integrates socially-aware information into the agent's navigation policy, ensuring safe and efficient navigation in dynamic scenes.


\section{METHODOLOGY}
 \label{chapter::meth_overview}
\begin{figure*}[t]
    \centering
    \includegraphics[width=0.9\linewidth]{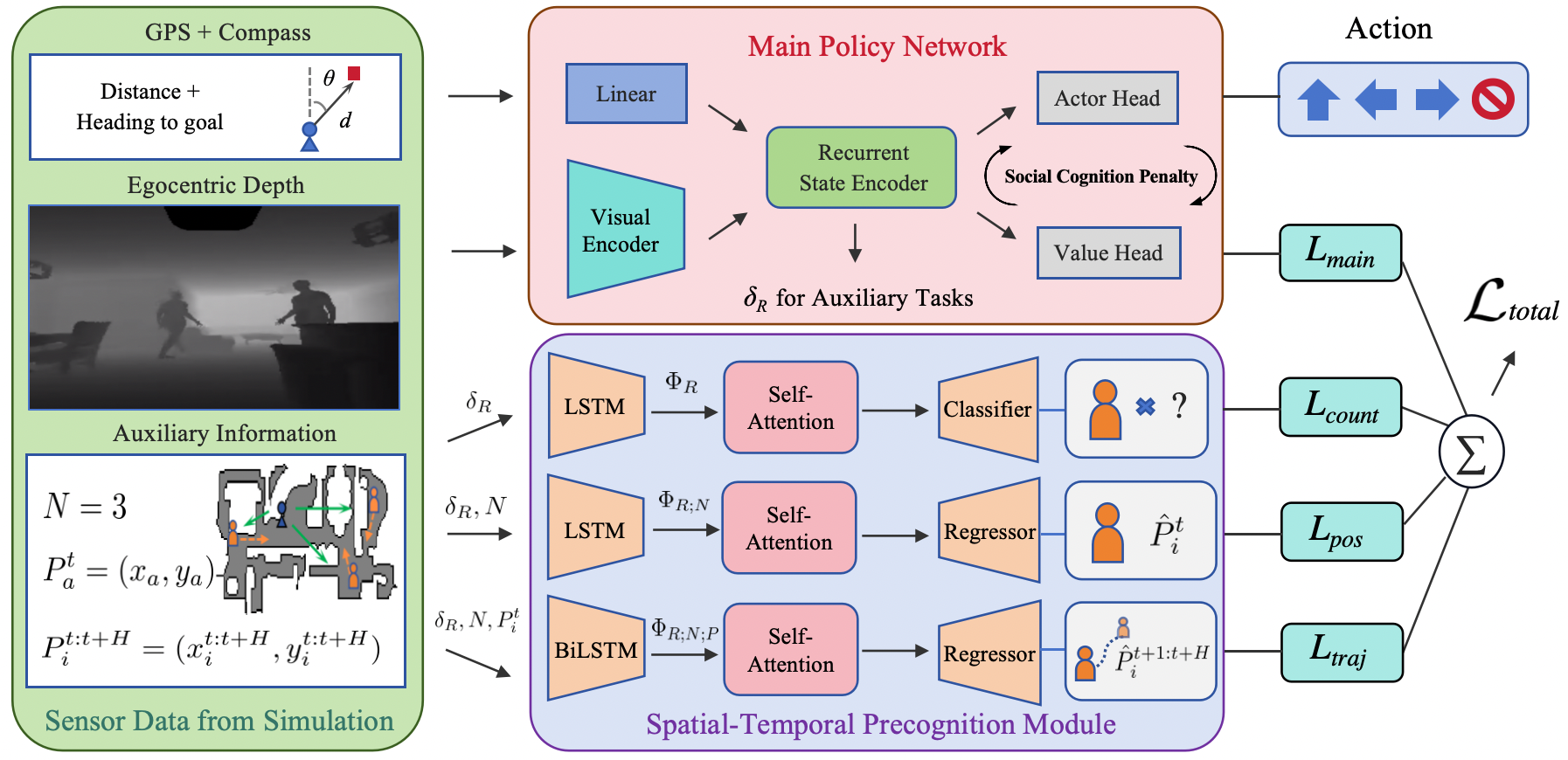}
    \caption{\textbf{\textit{Falcon} Overview}: The main policy network (top-right) takes Depth and GPS+Compass data as input. Its behavior is guided by the \textbf{Social Cognition Penalty}, which encourages socially compliant navigation and generates the main loss. During training, the output of the network's state encoder, combined with auxiliary information from the Habitat simulator, is processed by the \textbf{Spatial-Temporal Precognition Module} (bottom-right). Three socially-aware auxiliary tasks are then performed, producing auxiliary losses. The total loss is computed by weighting the main loss with the auxiliary losses.}
    \label{fig:meth_overview}
    \vspace{-2em}
\end{figure*}

\subsection{Problem Formulation}
Consider a social navigation task where a robot $a$ navigates in an environment populated by  \( N \) dynamic humans, \( i \in \{1, \dots, N\} \).
Starting from an initial configuration \( q_a \in Q \), the robot aims to continuously select actions to generate a path $\tau_a$ toward its goal configuration \( g_a \in Q \) while avoiding collision with static obstacles and dynamic humans. The overall objective is formulated as follows:
\begin{equation}
\begin{aligned}
\tau_a &= \arg \min_{\tau \in \mathcal{T}} \left( c_a(\tau) + \lambda_a c_a^s(\tau, {\tau}_{1:N}) \right) \\
\text{s.t.} \quad & A_a(\tau_a) \notin C_{\text{obs}},\quad A_a(\tau_a) \cap A_i(\tau_i) = \emptyset, \\
& \tau_a(0) = q_a, \quad \tau_a(T) = g_a
\end{aligned}
\end{equation}

where $c_a$ is the path cost guiding the robot to its goal; $c_a^s$ is the cost term accounts for social norms; 
$A(\tau)$ represents the volume occupied along trajectory $\tau$; 
\( C_{obs} \) represents static obstacles; $T$ is the episode end time; 
and \( \lambda_a \) is a weight factor.
The constraints make sure the robot does not collide with static obstacles and humans before reaching its goal.

Fig.~\ref{fig:meth_overview} is an overview of \textit{Falcon}. 
The \textbf{Main Policy Network} takes a depth image and a point goal at each timestep as inputs and directly outputs the robot's actions for the next step. It is trained using PointNav rewards along with our proposed \textbf{Social Cognition Penalty (SCP)}.
In addition, the network is accompanied by the \textbf{Spatial-Temporal Precognition Module} that facilitates several auxiliary tasks during training.
We detail each module in the subsequent sections.

\subsection{Main Policy Network}

The main policy network has two key components: the state encoders that extract visual and temporal features from observations, and a social cognition penalty that encourages social compliance. 
At each timestep, the point goal is transformed by a linear encoder whereas the depth image is processed by a ResNet-50~\cite{resnet} visual encoder.
A 2-layer LSTM~\cite{lstm} extracts the temporal features, which are then input to the actor head to produce actions and a value head to predict rewards.
The LSTM hidden state is also used as a latent variable \(\delta_R\) to connect with the auxiliary task module (see the next section).
During training, the policy is guided by a reward function that encourages goal-reaching behaviors. 
Classic PointNav reward function~\cite{wijmans2019dd} is used at each timestep $t$:
\begin{equation}
R^{\hspace{1pt}t}_\text{pointnav} = -\beta_d\Delta_d - r_{\text{slack}} + \beta_{\text{succ}} \cdot I_{\text{succ}}
\end{equation}
where \( \Delta_d \) is the change in geodesic distance to the goal, \( r_{\text{slack}}\) is a step penalty to prevent unnecessary actions,   $t$ denotes the current time step, and \( I_{\text{succ}} \) is an indicator for successful navigation, $\beta_d$ and \( \beta_{\text{succ}} \) are weight terms.

However, PointNav's reward function is inadequate for SocialNav as it ignores dynamic environments and social interactions.
To address this, we introduce \textbf{Social Cognition Penalty (SCP)}, a set of penalties designed to promote adherence to social norms.
Specifically, it contains: 

\mypara{Obstacle Collision Penalty.}
Collisions with static obstacles or humans are penalized by 

\begin{equation}
    r_{\text{coll}} = \beta_s \cdot I_{\text{s\_coll}} + \beta_h \cdot I_{\text{h\_coll}}
\end{equation}
where \( I_{\text{s\_coll}} \) and \( I_{\text{h\_coll}} \) are indicator variables representing collision with static obstacles and humans, respectively, and \( \beta_s\) and \( \beta_h\) are the corresponding penalty weights.

\mypara{Human Proximity Penalty.}
This penalty ensures the agent maintains a safe distance from humans and is given by 
\begin{equation}
r_{\text{prox}} = \sum_{i=1}^{N} 
\begin{cases} 
\beta_{\text{prox}} \cdot \exp\left( -d_i^t\right) & \text{if } d_i^t < 2.0 \mathrm{~m} \\
0 & \text{if } d_i^t \geq 2.0 \mathrm{~m} 
\end{cases}
\end{equation}
here \( d_i^t = \lVert \tau_a(t) - \tau_i(t)  \rVert \) is robot's distance to the \(i\)-th human at time step t. As \( d_i \) decreases, the penalty increases exponentially to prompt the agent to avoid humans. The penalty is removed when the robot is within 2.0 m to goal.

\mypara{Trajectory Obstruction Penalty.}
This penalty discourages the robot from obstructing future $H$-step human trajectories. It anticipates potential obstructions by considering both current and future positions, with earlier overlaps penalized more heavily. This penalty is calculated as:

\begin{equation}
r_{\text{traj}} = \sum_{k=t+1}^{t+H} \sum_{i=1}^{N} 
\begin{cases} 
\beta_{\text{traj}} \cdot \left( \frac{1}{k - t + 1} \right) & \text{if } d_{\text{traj}\_i}^k < 0.05 \mathrm{~m} \\
0 & \text{if } d_{\text{traj}\_i}^k \geq 0.05 \mathrm{~m} 
\end{cases}
\end{equation}
where \( d_{\text{traj}\_i}^k = \lVert \tau_a(k) - \tilde{\tau}_i(k) \rVert \) is the distance between the robot’s and human's future trajectories at the $k$-th time step. This penalty is also canceled when the robot nears its goal within 2.0 m.

\textbf{The overall reward function} is the combination of the goal-directed reward $R^{\hspace{1pt}t}_\text{pointnav}$ and the SCP penalties $R^{\hspace{1pt}t}_\text{scp}$:
\begin{equation}
R^{\hspace{1pt}t}_\text{socialnav} = R^{\hspace{1pt}t}_\text{pointnav} - R^{\hspace{1pt}t}_\text{scp}
\end{equation}
where $R^{\hspace{1pt}t}_\text{scp}$ is defined as:
\begin{equation}
 R^{\hspace{1pt}t}_\text{scp} = r_{\text{coll}} + r_{\text{prox}} + r_{\text{traj}}
\end{equation}
The main policy network is trained using DD-PPO~\cite{wijmans2019dd} with the PPO loss \( \mathcal{L}_{\text{main}} \).

\subsection{Spatial-Temporal Precognition Module}

This module utilizes three socially-aware auxiliary tasks to boost the robot's grasp of spatial-temporal dynamics.
As shown in Fig.~\ref{fig:meth_overview}, similar networks with an LSTM encoder and a self-attention block are used for each auxiliary task. 
We discuss each of these tasks as follows.

\mypara{Human Count Estimation.} 
This task aims to estimate the overall number of humans and the output is a discrete value between 0 and \(M\) (\(M = 6\) in our experiments). 
The LSTM takes as input the latent variable \(\delta_R\) from the main policy network, and outputs an encoded \(\Phi_R\), which is then used for the self-attention layer~\cite{vaswani2017attention} with \(Q = K = V = \Phi_R\). 
The classifier \(\phi_{\text{count}}\) then predicts the probabilities of human counts \(\hat{n}_k\), using the attention output $\mathbf{A}^t$ at time step $t$ :
\begin{equation} 
\hat{n}_k = \phi_{\text{count}}(\mathbf{A}^{t}) \quad \text{for } k \in \{0, 1, \dots, M\}
\end{equation} 
The loss is computed using Cross-Entropy:
\begin{equation} 
\mathcal{L}_{\text{count}} = - \sum_{k=0}^{M} n_k \log(\hat{n}_k)
\end{equation} 
where \(n_k\) is the indicator for the true count, and \(\hat{n}_k\) is the predicted probability for \(k\) humans.

\mypara{Current Position Tracking.} 
This task tracks the 2D locations of humans relative to the robot. 
Inputs are \(\delta_R\) and the oracle number of human agents \(N\) in the scene. 
The LSTM output \(\Phi_{R;N}\) is used as \(Q = K = V = \Phi_{R;N}\) in the self-attention layer. The regressor \(\phi_{\text{pos}}\) predicts each human's position \(\hat{\mathbf{P}}_{i}^{\hspace{1pt}t}\) at time \(t\):
\begin{equation}
    \hat{\mathbf{P}}_{i}^{\hspace{1pt}t} = \phi_{\text{pos}}(\mathbf{A}^{t})
\end{equation}
We use Mean Squared Error (MSE) between the predicted and true positions, with a mask \(\mathcal{M}\) to handle when there are less than $M$ humans in the scene:
\begin{equation}
    \mathcal{L}_{\text{pos}} = \frac{1}{|\mathcal{M}|} \sum_{i \in \mathcal{M}} \|\hat{\mathbf{P}}_{i}^{\hspace{1pt}t} - \mathbf{P}_{i}^{\hspace{1pt}t}\|^2
\end{equation}

\mypara{Future Trajectory Forecasting.} 
This task predicts human trajectories over multiple time steps. It uses a bi-directional stacked LSTM due to its complexity. Inputs include \(\delta_R\), \(N\), and current human positions \(P^t_i\). The LSTM output \(\Phi_{R;N;P}\) is used as \(Q = K = V = \Phi_{R;N;P}\) in the attention layer. The regressor \(\phi_{\text{traj}}\) predicts future trajectories over \(H\) steps:
\begin{equation}
\hat{\mathbf{P}}_{i}^{\hspace{1pt}t+1:t+H} = \phi_{\text{traj}}(\mathbf{A}^{t}).
\end{equation}
Similarly, MSE is applied with the mask \(\mathcal{M}\):
\begin{equation}
\mathcal{L}_{\text{traj}} = \frac{1}{|\mathcal{M}|} \sum_{i \in \mathcal{M}} \|\hat{\mathbf{P}}_{i}^{\hspace{1pt}t+1:t+H} - \mathbf{P}_{i}^{\hspace{1pt}t+1:t+H}\|^2
\end{equation}
The auxiliary loss is defined as
\begin{equation}
\mathcal{L}_{\text{aux}} = \mathcal{L}_{\text{count}} + \mathcal{L}_{\text{pos}} + \mathcal{L}_{\text{traj}}
\end{equation}
During training, the main policy network and these auxiliary tasks are optimized together.
The total loss function is a weighted sum of the main policy loss \( \mathcal{L}_{\text{main}} \) and the auxiliary loss \( \mathcal{L}_{\text{aux}} \):
\begin{equation}
   \mathcal{L}_{\text{total}} =\beta_{\text{main}} \mathcal{L}_{\text{main}} + \beta_{\text{aux}} \mathcal{L}_{\text{aux}}
\end{equation}
where $\beta_{\text{main}}$ and $\beta_{\text{aux}}$ are their respective loss weights.

\section{EXPERIMENTS}
\label{chapter：：Experiments}
In this section, we first introduce the proposed benchmark and its two datasets, followed by the experiment setup and evaluation metrics. 
Next, we show the results of \textit{Falcon} against prior methods. 
Finally, we conduct ablation analysis of two key components,  Social Cognition Penalty and Spatial-Temporal Precognition Module.

\subsection{Datasets}
\label{chapter::datasets}

While existing datasets for SocialNav offer valuable environments, many of them~\cite{igibsonchallenge2021,makoviychuk2021isaac} lack diversity in scene types. 
Other datasets~\cite{vuong2023habicrowd,cancelli2023exploiting}, though rich in scenes, do not adequately balance human density and the human movement patterns are unnatural or do not have animation at all.
To address these limitations, we introduce a benchmark with \textbf{Social-HM3D} and \textbf{Social-MP3D}, two simulation datasets derived from photo-realistic HM3D~\cite{ramakrishnan2021habitat} and MP3D~\cite{chang2017matterport3d}. 
The statistical properties of our benchmark, in comparison with existing datasets, are shown in Table~\ref{tab::dataset_stats}. Our datasets offer a wide variety of environments with carefully calibrated human density, incorporating realistic human motions and natural movement patterns. 
These features ensure balanced interaction dynamics across diverse scenes, facilitating the development of more effective social navigation algorithms.
We describe these key designs of our benchmark below:

\begin{table}[t]
 \resizebox{1\columnwidth}{!}{
 \tablestyle{2pt}{1.1}
\begin{tabular}{x{60}x{30}x{65}x{32}x{30}}
\toprule
  Dataset & \begin{tabular}[c]{@{}c@{}}Num.\\Scenes\end{tabular}  & \begin{tabular}[c]{@{}c@{}}Scene\\Type\end{tabular}    & \begin{tabular}[c]{@{}c@{}}Max Num.\\Humans\end{tabular} & \begin{tabular}[c]{@{}c@{}}Natural \\Motions \end{tabular}  \\
\midrule
iGibson-SN~\cite{igibsonchallenge2021} & 15 & residence & 3 & \xmark \\
\arrayrulecolor[gray]{0.8} 
\hline
Isaac Sim~\cite{makoviychuk2021isaac} & 7 & \begin{tabular}[c]{@{}c@{}}residence, office, \\ depot, etc.\end{tabular} & 7 & \cmark\\
\hline
HabiCrowd~\cite{vuong2023habicrowd} & 480 & \begin{tabular}[c]{@{}c@{}}residence, office, \\ gym, etc.\end{tabular} & 40 & \xmark\\
\hline
HM3D-S~\cite{cancelli2023exploiting} & 900 & \begin{tabular}[c]{@{}c@{}}residence, office, \\ shop, etc.\end{tabular}  & 3 & \xmark\\
\arrayrulecolor{black} 
\hline
\textbf{Social-HM3D} & 844 & \begin{tabular}[c]{@{}c@{}}residence, office, \\ shop, etc.\end{tabular} & 6 & \cmark\\
\arrayrulecolor[gray]{0.8} 
\hline
\textbf{Social-MP3D} & 72 & \begin{tabular}[c]{@{}c@{}}residence, office, \\ gym, etc.\end{tabular} & 6 & \cmark \\
\arrayrulecolor{black} 
\bottomrule
\end{tabular}
}
\caption{Statistics Comparison of SocialNav Datasets/Simulators: Our proposed Social-HM3D and Social-MP3D datasets feature extensive scene diversity and realistic interaction design, addressing the shortcomings of previous datasets which often relied on oversimplified human behaviors and imbalanced interaction dynamics.}
\label{tab::dataset_stats}
\vspace{-1em}
\end{table}
\begin{figure}[t]
    \centering
    \includegraphics[width=0.8\linewidth]{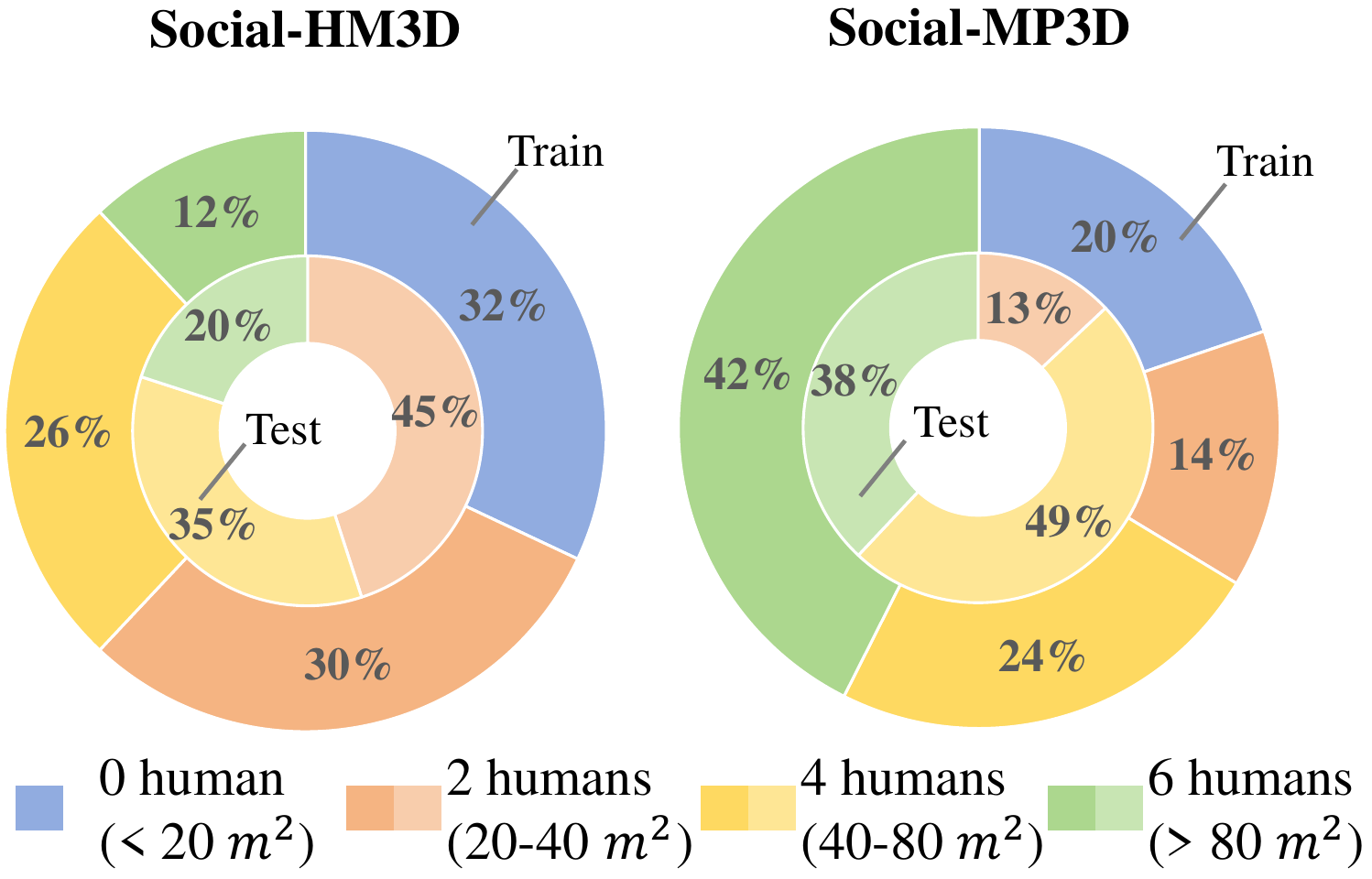}
    \caption{Human Distribution by Scene Area in Social-HM3D and Social-MP3D (Train/Test): Our benchmark balances social density for human-robot interactions while avoiding overcrowding.}
    \label{fig::dataset_pie}
    \vspace{-2em}
\end{figure}
\mypara{Realistic Agent Trajectories.} 
Unlike previous datasets that leverage random walks or repetitive movements~\cite{igibsonchallenge2021,vuong2023habicrowd}, our dataset provides more realistic and reasonable human trajectories with goals.
Each human is assigned two goals within the robot's navigation area, intended to encourage broader movement across the environment and increase the chance to interact with the robot. 

\mypara{Natural Human Behaviors.} 
Our dataset realistically simulates human dynamics, with humans alternating between movement and rest, stopping once their tasks are complete. Walking speeds are randomized between 0.8 to 1.2 times the robot's speed, reflecting natural human movement. Humans react and avoid each other using ORCA algorithm~\cite{van2011reciprocal}. Realistic walking animations for both humans and robots enhance the visual quality and authenticity of social interactions.

\mypara{Reasonable Human Density.} 
Our datasets group humans by scene area size to balance social interaction density. 
Fig.~\ref{fig::dataset_pie} shows the distribution of human numbers in the datasets, which are carefully selected to ensure meaningful human-robot interactions while avoiding overcrowding that could hinder the movement of both the robot and humans.

\mypara{Diversity and Scalability}
As shown in Table \ref{tab::dataset_stats}, our datasets contain a broad range of environments, featuring 844 scenes derived from Social-HM3D and 72 scenes in Social-MP3D.
The diversity of scenes and the inclusion of active humans at carefully chosen densities enhance its scalability to various tasks. 
Besides supporting SocialNav, our benchmark can be directly extended to social object/image navigation.

\subsection{Experiment Setup}
\label{chapter::exp_detail}
\textbf{Metrics.}
Our benchmark metrics build upon existing works~\cite{anderson2018evaluation,igibsonchallenge2021} and focus on two principal perspectives: task completion and adherence to SocialNav objectives. For task completion, we use Success Rate (Suc.), Success weighted by Path Length (SPL) and Success weighted by Time Length (STL).
For social norms, we evaluate Human-Robot Collision Rate (H-coll) and Personal Space Compliance (PSC). 
Considering the human collision radius is 0.3m and the robot is 0.25m, the PSC distance threshold is set to 1.0m.

\textbf{Baseline Models.}
We mainly compare our method with a recent state-of-the-art social navigation method, namely Proximity-Aware~\cite{cancelli2023exploiting}. It introduces two auxiliary tasks that model both the distance and direction of humans, effectively capturing their proximity at current.
We also implement two classic baseline methods, including A-star (A*)\cite{hart1968formal} and ORCA\cite{van2011reciprocal} for comparison. 
A* and ORCA are both rule-based path-planning algorithms. 
ORCA has oracle access to agent positions and velocities to dynamically adjust the planned route.
Both Proximity-Aware and our \textit{Falcon} are RL-based. 
All methods only take depth image as visual inputs to ensure a fair comparison.

\textbf{Implementation Details.}
We train the RL agents using DD-PPO algorithm \cite{wijmans2019dd} with identical hyperparameters. 
Each algorithm is run three times with different random seeds, and we report the mean and standard deviation of each metric.
Our model is initialized with pre-trained weights from a PointNav model~\cite{ramakrishnan2021habitat} and is fine-tuned for 10 million steps on the SocialNav task.
Training is conducted on 4 Nvidia RTX 3090 GPUs with 8 parallel environments.
Models are trained on the Social-HM3D train set and tested on both Social-HM3D and Social-MP3D, with Social-MP3D results used to evaluate \textit{Falcon}'s zero-shot generalization.

\subsection{Result Analysis}

\begin{table*}[ht]
\centering
\resizebox{1\textwidth}{!}{
\tablestyle{4pt}{1.1}
\begin{tabular}{x{45}x{38}y{68}x{45}x{45}x{45}x{45}x{45}}

\toprule

Dataset &\multicolumn{2}{c} {Method} & Suc. $\uparrow$ & SPL $\uparrow$ & STL $\uparrow$ & PSC $\uparrow$ & H-Coll $\downarrow$ \\

\midrule

\multirow{4}{*}{Social-HM3D} 
    & \multirow{2}{*}{Rule-Based} 
         & A$^*$~\cite{hart1968formal}                     & {\ul 46.14\scriptsize{$\pm0.7$}} & {\ul 46.14\scriptsize{$\pm0.7$}}    & {\ul 46.12\scriptsize{$\pm0.7$}}         & {\ul 90.56\scriptsize{$\pm0.2$}}           & 53.50\scriptsize{$\pm0.9$} \\
    &    & ORCA~\cite{van2011reciprocal}                   & 38.91\scriptsize{$\pm0.1$}      &  38.91\scriptsize{$\pm0.1$}        & 38.44\scriptsize{$\pm0.1$}              & 90.55\scriptsize{$\pm0.4$}                & 47.52\scriptsize{$\pm1.7$} \\
    \cmidrule(lr){2-8} 
    & \multirow{2}{*}{RL} 
         & Proximity-Aware~\cite{cancelli2023exploiting}   & 20.11\scriptsize{$\pm1.3$}      & 18.57\scriptsize{$\pm1.9$}         & 19.51\scriptsize{$\pm1.5$}                & \textbf{92.91}\scriptsize{$\pm0.5$}    & \textbf{33.99}\scriptsize{$\pm0.7$} \\ 
    &    & {\cellcolor{gray!25} Falcon}                    & {\cellcolor{gray!25} \textbf{55.15}}\scriptsize{$\pm 0.6$}  & {\cellcolor{gray!25} \textbf{55.15}}\scriptsize{$\pm0.7$}  & {\cellcolor{gray!25} \textbf{54.94}}\scriptsize{$\pm0.7$} & \cellcolor{gray!25} {89.56}\scriptsize{$\pm1.4$} & {\cellcolor{gray!25}{\ul 42.96\scriptsize{$\pm1.1$}}} \\
    
\midrule
\midrule

   \multirow{4}{*}{Social-MP3D} 
   & \multirow{2}{*}{Rule-Based} 
        & A$^*$~\cite{hart1968formal}                    & {\ul 43.85\scriptsize{$\pm0.3$}}    & {\ul 43.85\scriptsize{$\pm0.3$}}           & {\ul 43.85\scriptsize{$\pm0.3$}}               & 86.74\scriptsize{$\pm3.4$}              & 57.94\scriptsize{$\pm1.5$} \\
    &   & ORCA~\cite{van2011reciprocal}                  & 40.38\scriptsize{$\pm0.3$}     & 40.38\scriptsize{$\pm0.3$}            & 39.51\scriptsize{$\pm0.2$}               & {\ul 91.76\scriptsize{$\pm0.4$}}              & 47.16\scriptsize{$\pm0.2$} \\
    \cmidrule(lr){2-8} 
    & \multirow{2}{*}{RL} 
        & Proximity-Aware~\cite{cancelli2023exploiting}   & 18.45\scriptsize{$\pm1.4$}     & 17.09\scriptsize{$\pm2.8$}            & 16.41\scriptsize{$\pm1.5$}                & \textbf{93.37}\scriptsize{$\pm0.9$}    & \textbf{32.18}\scriptsize{$\pm3.3$} \\
    &    &  {\cellcolor{gray!25}  Falcon}                    & {\cellcolor{gray!25} \textbf{55.05}\scriptsize{$\pm0.7$}}  & {\cellcolor{gray!25} \textbf{55.04}\scriptsize{$\pm0.6$}} & {\cellcolor{gray!25} \textbf{54.80}\scriptsize{$\pm1.0$}} &  {\cellcolor{gray!25} 90.01\scriptsize{$\pm1.2$}}      & {\cellcolor{gray!25} {\ul 42.19\scriptsize{$\pm0.9$}}} \\
 \bottomrule
\end{tabular}

}
\caption{Performance Evaluation of SocialNav Tasks for Rule-Based and RL-Based Methods on Social-HM3D~(upper group) and Social-MP3D~(lower group). Data in the table represents percentages. 
We \textbf{bold} the best results and {\ul underline} the second best results. }
\label{tab::main_table}
\vspace{-1em}
\end{table*}

We conduct experiments to investigate several key aspects: 

\begin{itemize}
    \item Effectiveness of our algorithm against prior methods.
    \item Impacts of our auxiliary tasks.
    \item Individual and cooperative effects of the Social Cognition Penalty~(SCP) and Spatial-Temporal Precognition Module~(SPM).
\end{itemize}

Table~\ref{tab::main_table} presents the comparison results on the Social-HM3D test set and zero-shot results on Social-MP3D test set. 
As we can see, the superior results demonstrate the efficiency of \textit{Falcon} in both goal-reaching and social compliance, while the comparable zero-shot results highlight its generalization to unseen environments. 
Qualitatively, Fig.~\ref{fig:comp_all} provides examples illustrating various classic encounters in SocialNav tasks, where our method outperforms the others. 
Based on these results, we derive the following findings:

\begin{figure}[t]
    \centering
    \begin{subfigure}{\linewidth}
        \centering
        \includegraphics[width=\linewidth]{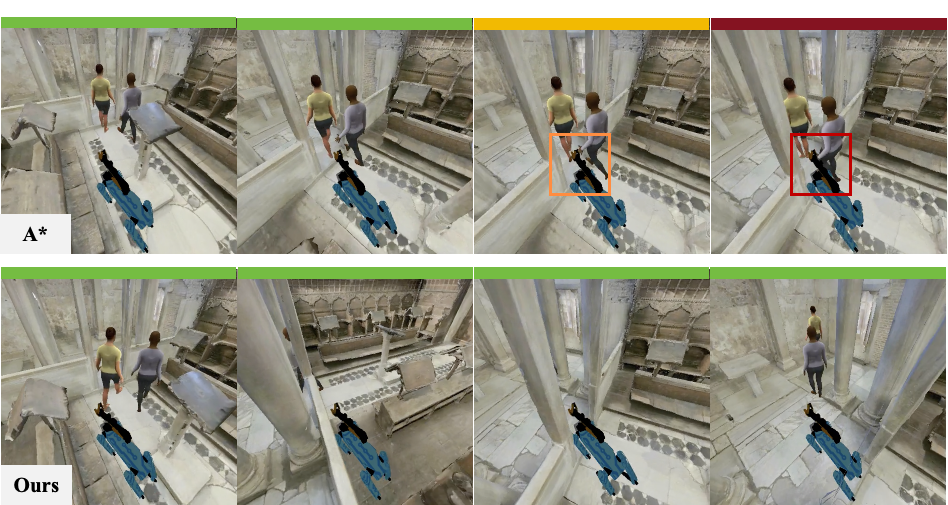}
        \caption{Person Following: A* algorithm does not wait for humans and causes a collision, while our method successfully follows.}
        \label{fig:comp_astar}
    \end{subfigure}
    
    \vspace{0.1cm}  
    
    \begin{subfigure}{\linewidth}
        \centering
        \includegraphics[width=\linewidth]{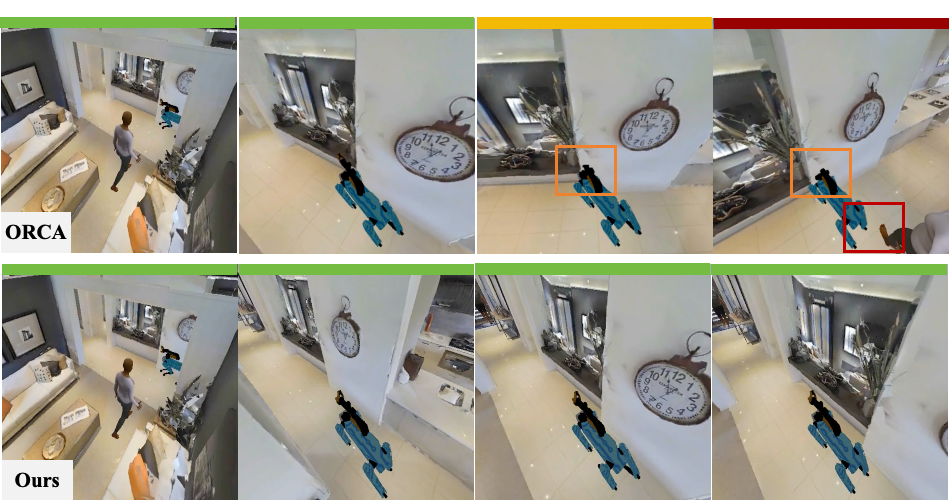}
        \caption{Intersection Encounter: ORCA algorithm fails to avoid a person and collides with a static obstacle, instead our method passes safely.}
        \label{fig:comp_orca}
    \end{subfigure}
    
    \vspace{0.1cm}  
    
    \begin{subfigure}{\linewidth}
        \centering
        \includegraphics[width=\linewidth]{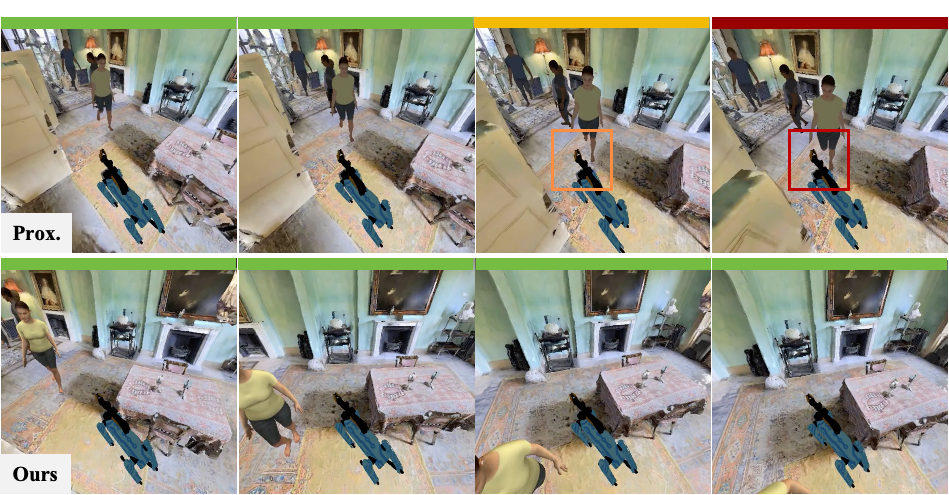}
        \caption{Frontal Approach: The Proximity-Aware method collides when crossing in front directly, while our method avoids safely by anticipating human path.}
        \label{fig:comp_prox}
    \end{subfigure}
    
    \caption{Comparisons of SocialNav Algorithms in Different Encounters: Our method outperforms other algorithms across various encounters. Green indicates safe behaviors, orange indicates risky behaviors (e.g., proximity to humans or collisions with obstacles), and red indicates unsafe behaviors (i.e., collisions with humans).}
    \label{fig:comp_all}
\vspace{-2em}
\end{figure}

\mypara{Finding 1: Future-aware methods are more efficient and safer than static and situation-aware approaches.}
Static path-planning algorithms like A* determine a fixed route and cannot adapt to dynamic environments, causing collision with humans (See Fig.\ref{fig:comp_astar}). 
In contrast, situation-aware obstacle-avoidance approaches, such as ORCA and Proximity-Aware, react to the current environment by adjusting paths to avoid humans and obstacles. 
However, these methods have limitations: rerouting often takes time, and delayed reactions increase collision risks. 
ORCA assumes unrestricted movement and causes a collision with static obstacles (See Fig.\ref{fig:comp_orca}). 
Similarly, the Proximity-Aware method struggles due to its inability to predict human movement, leading to failed short-term adjustments (See Fig.\ref{fig:comp_prox}). 
\textit{Falcon} is able to proactively adjust to dynamic human movements and reach the goal location effectively.

\begin{figure}[t]
    \centering
    \includegraphics[width=0.9\linewidth]{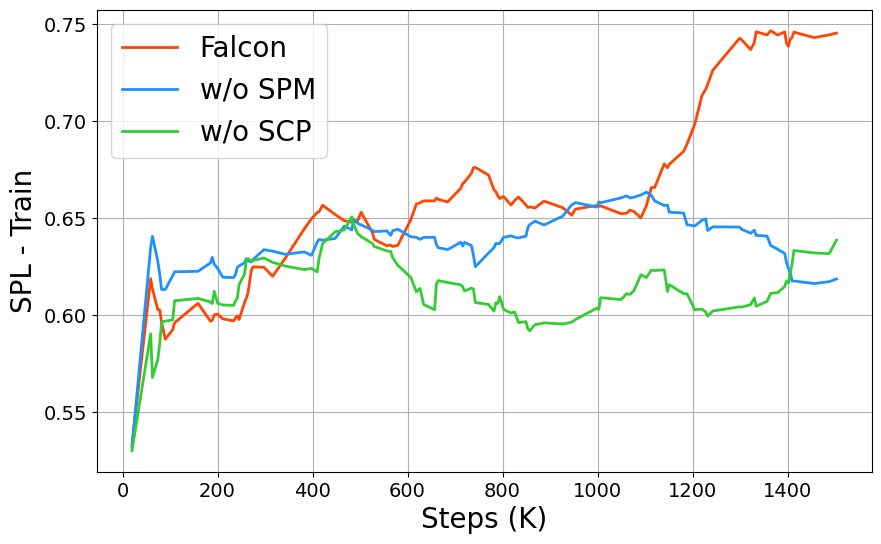}
    \caption{Training Curve of SPL for Ablation Study: The full \textit{Falcon} model with SPM and SCP converges faster and performs better.}
    \label{fig::ablation_curve}
    \vspace{-2em}
\end{figure}

\begin{table}[t]
 \resizebox{1\columnwidth}{!}{
 \tablestyle{4pt}{1.1}
\begin{tabular}{z{16}z{16}z{16}z{16}|y{18}x{18}x{18}x{18}x{29}}
\toprule
       \begin{tabular}[c]{@{}c@{}}SPM.\\ Count\end{tabular} & \begin{tabular}[c]{@{}c@{}}SPM.\\ Pos\end{tabular} & \begin{tabular}[c]{@{}c@{}}SPM.\\ Traj\end{tabular} & SCP & Suc.~$\uparrow$ & SPL~$\uparrow$ & STL~$\uparrow$ & PSC~$\uparrow$ & H-Coll~$\downarrow$ \\
      \midrule
    \multicolumn{4}{c|}{\fadedtext{PointNav~(w/o Aux. Task)}} & \fadedtext{40.94} & \fadedtext{34.14} & \fadedtext{11.50} & \fadedtext{\textbf{90.82}} & \fadedtext{53.54} \\ 
    \midrule
       $\checkmark$ &  & & &51.43 & 51.42 & 51.16 & 90.53 & 46.46 \\
       & $\checkmark$ & & & 53.17 & 53.17 & 52.95 & 90.06 & 44.07 \\
      & & $\checkmark$ & & 54.00 & 53.99 & 53.92 & 89.46 & 43.88 \\
                 & & &   $\checkmark$  &51.24 & 51.24 & 51.08 & 90.41 & 48.11 \\
      $\checkmark$ & $\checkmark$ & $\checkmark$ & & 53.63 & 53.63 & 53.40 & 89.33 & 44.89 \\
    {\cellcolor{gray!25} $\checkmark$} & {\cellcolor{gray!25} $\checkmark$} & {\cellcolor{gray!25} $\checkmark$} & {\cellcolor{gray!25} $\checkmark$} & {\cellcolor{gray!25} \textbf{55.15}} & {\cellcolor{gray!25} \textbf{55.15}} & {\cellcolor{gray!25} \textbf{54.94}} & {\cellcolor{gray!25} 89.56} & {\cellcolor{gray!25} \textbf{42.96}} \\
    
    \bottomrule
\end{tabular}
}
\caption{Ablation Study for \textit{Falcon}. The model trained solely with the PointNav algorithm~\cite{ramakrishnan2021habitat} serves as the baseline. SPM.Count, SPM.Pos, and SPM.Traj refer to three auxiliary tasks: Humanoid Count Estimation, Current Position Tracking, and Future Trajectory Forecasting. Data in the table are percentages.} 
\label{tab::ablation_study}
\vspace{-0.5em}
\end{table}

\mypara{Finding 2: Auxiliary tasks contribute to performance improvement, with trajectory prediction playing the most significant role.}
Table~\ref{tab::ablation_study} presents experiments with various auxiliary task choices. Each task individually improves navigation performance compared to the PointNav baseline~\cite{ramakrishnan2021habitat}.
Among them, trajectory forecasting (SPM.Traj) proves to be the most effective auxiliary task, with the success rate notably increasing from 40.94\% to 54.00\%. 
This highlights the value of explicit trajectory prediction in SocialNav. 

\mypara{Finding 3: SCP coordinates and complements SPM, leading to a significant improvement and faster training.} 
Table~\ref{tab::ablation_study} shows that SCP plays a critical role in enhancing the model's performance, particularly when integrated with the SPM. 
As we can see, the SPM with its three auxiliary tasks (Count, Pos, and Traj) does not improve much upon training them individually (SPM's 53.63\% vs. SPM.Pos's 53.17\% and SPM.Traj's 54.00\%).
With SCP, the full system achieves significant improvement over SPM (55.15\% vs. 53.63\%).
Also, as shown in Fig.~\ref{fig::ablation_curve}, the model trained with both SPM and SCP exhibits faster convergence during training (evident before 1400K steps). 
These results show that SPM tasks cannot be effectively integrated without the guidance of SCP, which helps the model balance tasks and better leverage available information.

\mypara{Limitations.}
\textit{Falcon} can achieves high success, while Proximity-Aware, despite low success ($\sim$20\%), excels in avoiding collisions. 
This reveals a limitation in existing metrics, which may prioritize social comfort over task success in crowded environments.
Also, our current benchmark does not involve higher-level human behaviors like yielding.

\section{CONCLUSIONS}

In this paper, we introduce a novel SocialNav benchmark with two datasets, Social-HM3D and Social-MP3D.
We propose \textit{Falcon}, a future-aware  method for social navigation in realistic human-populated scenes.
Our method achieves state-of-the-art results over rule-based and recent reinforcement learning models on our proposed new benchmark.
We believe our benchmark, together with our method, will facilitate future research and applications on social navigation.




\clearpage

\bibliographystyle{IEEEtran} 
\bibliography{IEEEexample}

\end{document}